\documentclass[sigconf,balance]{acmart}
\usepackage{multirow}
\usepackage{enumitem}

\makeatletter
\newbox\sTwoDBibCitationBox
\AtBeginMaketitle{%
  \g@addto@macro\@printtopmatter{\vspace*{4pt}}%
  \let\sTwoDOriginalBibCitation\@mkbibcitation
  \def\@mkbibcitation{%
    \setbox\sTwoDBibCitationBox=\vbox{%
      \hsize=\columnwidth
      \vskip24pt
      \sTwoDOriginalBibCitation}%
    \noindent\box\sTwoDBibCitationBox}}
\makeatother

\skip\footinscopyrightpermission=9pt plus 2pt minus 1pt

\AtBeginDocument{%
  }

\copyrightyear{2026}
\acmYear{2026}
\setcopyright{cc}
\setcctype{by}
\acmConference[MM '26]{Proceedings of the 34th ACM International Conference on Multimedia}{November 10--14, 2026}{Rio de Janeiro, Brazil}
\acmBooktitle{Proceedings of the 34th ACM International Conference on Multimedia (MM '26), November 10--14, 2026, Rio de Janeiro, Brazil}
\acmDOI{10.1145/3767308.3836562}
\acmISBN{979-8-4007-2213-4/2026/11}

\begin{document}
\raggedbottom

\title[
  \parbox{0.72\textwidth}{
    From Semantic Grounding to Decision Optimization:\newline
    A Unified Framework for Long-Horizon UAV Vision-Language Navigation
  }
]{
  From Semantic Grounding to Decision Optimization:
  A Unified Framework for Long-Horizon UAV Vision-Language Navigation
}

\author{Zeyuan Ma}
\orcid{0009-0007-6494-2127}
\affiliation{%
  \institution{State Key Laboratory of Virtual Reality Technology and Systems, Beihang University}
  \city{Beijing}
  \country{China}}
\affiliation{%
  \institution{School of Computer Science and Engineering, Beihang University}
  \city{Beijing}
  \country{China}}
\email{zeyuanma@buaa.edu.cn}

\author{Jiaxin Chen}
\orcid{0000-0002-0112-4166}
\authornote{Corresponding author}
\affiliation{%
  \institution{State Key Laboratory of Virtual Reality Technology and Systems, Beihang University}
  \city{Beijing}
  \country{China}}
\affiliation{%
  \institution{School of Computer Science and Engineering, Beihang University}
  \city{Beijing}
  \country{China}}
\email{jiaxinchen@buaa.edu.cn}

\author{Di Huang}
\orcid{0000-0002-2412-9330}
\affiliation{%
  \institution{School of Computer Science and Engineering, Beihang University}
  \city{Beijing}
  \country{China}}
\email{dhuang@buaa.edu.cn}
\renewcommand{\shortauthors}{Ma et al.}

\begin{abstract}
UAV vision-language navigation (UAV-VLN) focuses on enabling an aerial agent to follow natural-language instructions in open 3D environments from egocentric visual observations. Current approaches suffer from three coupled issues: weak grounding of instruction-relevant landmarks in visual observations, insufficient exploration of long-horizon history, and unstable decisions under local traps or repeated exploration. To address these issues, we propose a unified semantic-to-decision framework. First, we present an instruction-grounded semantic enhancement module that injects object-level semantics and relative spatial cues into the current observation state. Subsequently, we develop a relevance-aware dynamic temporal aggregation strategy that reweights the full history buffer while converting a few high-relevance frames into structured landmark prompts for the decoder. Finally, we devise a topology-aware decision method that combines local-optimum cognition with group-relative policy optimization under progress, goal, semantic, and path-compliance rewards. Experiments on the widely used AerialVLN and OpenFly benchmarks clearly demonstrate that our method achieves state-of-the-art performance. Our code is available at \url{https://github.com/mksasx/S2D-UAV-VLN}.
\end{abstract}

% \begin{CCSXML}
% <ccs2012>
%    <concept>
%        <concept_id>10010147.10010178.10010219</concept_id>
%        <concept_desc>Computing methodologies~Artificial intelligence</concept_desc>
%        <concept_significance>300</concept_significance>
%    </concept>
%    <concept>
%        <concept_id>10010147.10010178.10010224</concept_id>
%        <concept_desc>Computing methodologies~Computer vision</concept_desc>
%        <concept_significance>100</concept_significance>
%    </concept>
%    <concept>
%        <concept_id>10010147.10010178.10010205</concept_id>
%        <concept_desc>Computing methodologies~Natural language processing</concept_desc>
%        <concept_significance>500</concept_significance>
%    </concept>
% </ccs2012>
% \end{CCSXML}


% \ccsdesc[300]{Computing methodologies~Artificial intelligence}
% \ccsdesc[100]{Computing methodologies~Computer vision}
% \ccsdesc[500]{Computing methodologies~Natural language processing}


\maketitle

% Keep float, caption, and display separation compact but nonnegative.  This
% avoids stretched white bands without changing ACM fonts, margins, or columns.
\setlength{\textfloatsep}{4pt}
\setlength{\dbltextfloatsep}{4pt}
\setlength{\floatsep}{4pt}
\captionsetup{skip=3pt}
\setlength{\abovedisplayskip}{2pt}
\setlength{\belowdisplayskip}{2pt}
\setlength{\abovedisplayshortskip}{0pt}
\setlength{\belowdisplayshortskip}{2pt}
\setlength{\jot}{1.5pt}

\section{Introduction}
\begin{figure*}[t]
    \centering
    \includegraphics[width=0.90\textwidth]{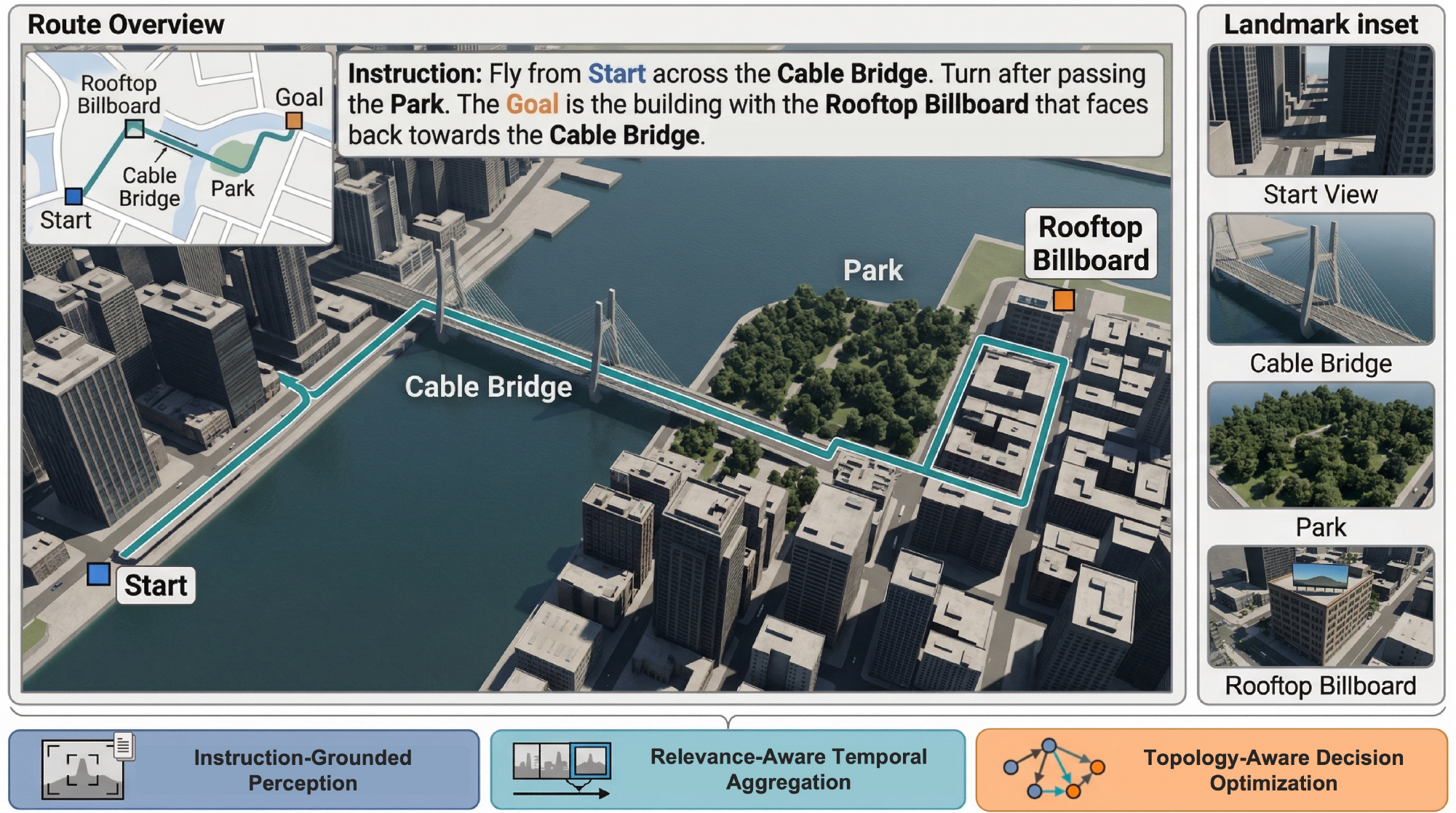}
    \caption{Illustration of the long-horizon UAV-VLN task. The example route shows that successful navigation depends on accurate landmark grounding, relevance-aware history selection, and stable topology-aware decisions.}
    \Description{A benchmark-style aerial route illustration for UAV vision-language navigation. An instruction at the top refers to landmarks such as a cable bridge, park, and rooftop billboard. The route runs from Start to Goal, and the bottom row highlights instruction-grounded perception, relevance-aware temporal aggregation, and topology-aware decision optimization.}
    \label{fig:task}
\end{figure*}

Vision-language navigation (VLN) studies how an embodied agent follows free-form language under partial visual observations~\cite{anderson2018vision}. For unmanned aerial vehicles (UAVs), this task is far more challenging, since the agent must reason across open 3D environments with drastic scale variations, sparse yet decisive landmarks, and trajectories much longer than in indoor or ground settings. This capability is important for inspection, search and rescue, logistics, and urban surveillance, where natural language provides a flexible interface for non-expert operators, although the present work evaluates simulation-based high-level navigation rather than real-world UAV deployment. Recent aerial research spans benchmark-style UAV-VLN datasets, high-fidelity simulation platforms, large-scale synthetic toolchains,  city-scale real-world aerial navigation corpora, low-altitude agentic application systems, and broader studies on aerial embodied-agents~\cite{liu2023aerialvln,wang2025openuav,gao2025openfly,lee2025citynav,lin2025openvln,zhao2025aerialagent,sautenkov2025uavvla,tian2025uavsmeetllms,yao2025aeroversereview}.

These resources have enabled measurable progress, with recent VLM/MLLM-based systems (\emph{e.g.}, CityNavAgent, FlightGPT, GeoNav) and contemporary open-world, end-to-end, or structured-view aerial variants (\emph{e.g.}, OpenVLN, UAV-VLN, grid-based aerial VLN) all demonstrating the potential of stronger multimodal reasoning for long-horizon aerial navigation~\cite{zhang2025citynavagent,cai2025flightgpt,xu2026geonav,lin2025openvln,saxena2025uavvln,zhao2025gridview}. Despite these advances, long-horizon UAV-VLN still suffers from a coupled failure chain. Current-view grounding is fragile because aerial landmarks are often small, repeated, and visually ambiguous, causing global image features to overlook the object phrases that directly determines the next action. Once grounding drifts, historical frame aggregation becomes noisy, as many stored frames cease to be relevant to the active sub-goal. The resulting temporal state destabilizes sequential decisions, incurring repeated exploration, semantically plausible yet incorrect stopping, or local looping in visually similar regions.

These errors do not stem from three isolated modules, but reflect a single core problem of state construction. Unlike recent aerial agents that primarily strengthen planner-side reasoning or coarse-to-fine geospatial search~\cite{cai2025flightgpt,xu2026geonav}, we focus on the policy state itself: grounding quality determines which past observations remain useful, temporal relevance controls whether the policy receives a clean state, and state quality governs the stability of topology-aware decision-making. We accordingly formulate aerial navigation as a progressive semantic-to-decision pipeline. As displayed in Fig.~\ref{fig:framework}, our framework first builds instruction-grounded observation tokens from landmark semantics, relative spatial cues, and local topology; then selectively reuses historical evidence through relevance-aware dynamic temporal aggregation; and finally refines action selection via topology-aware local-optimum recovery and GRPO-based policy optimization. Figure~\ref{fig:task} illustrates the task abstraction that motivates this design.

The main contributions of this work are summarized as follows:
\begin{itemize}[leftmargin=1.35em,topsep=1pt,itemsep=1pt,parsep=0pt,partopsep=0pt]
    \item We present a unified system-level semantic-to-decision framework for long-horizon UAV-VLN that jointly couples instruction-grounded current-view perception, relevance-aware temporal aggregation, and topology-aware action refinement.
    \item We develop two distinctive and complementary mechanisms, namely dynamic temporal aggregation (DTA) and local-optimum cognition (LOC) respectively, where DTA combines instruction-conditioned full-history weighting with sparse landmark prompts, and LOC detects topology-level stagnation and conditionally injects a semantically matched frontier cue. Together with composite trajectory rewards, they improve decision stability and robustness in long-horizon navigation.
    \item We validate the proposed framework on AerialVLN and OpenFly, where the results clearly indicate the effectiveness of the coupled design of semantic grounding, temporal aggregation, and topology-aware decision optimization.
\end{itemize}

\section{Related Work}
\subsection{Vision-Language Navigation}
VLN originated in indoor and ground settings, where agents navigate with denser local cues and shorter trajectories~\cite{anderson2018vision,fried2018speaker,krantz2020vlnce,hao2020hamt}. UAV-VLN is harder because the agent must reason over open 3D environments, severe scale changes, sparse landmarks, and much longer instruction horizons. The aerial benchmark ecosystem has expanded in several directions: AerialVLN and OpenFly provide benchmark-style long-horizon evaluation, OpenUAV introduces a more realistic simulation platform together with the UAV-Need-Help benchmark, and CityNav targets real-world city-scale aerial navigation~\cite{liu2023aerialvln,gao2025openfly,wang2025openuav,lee2025citynav}. Recent studies further broaden this space toward open-world aerial VLN, end-to-end UAV-VLN pipelines, delivery-oriented low-altitude navigation, dual-agent aerial VLN, and general aerial embodied-agent systems~\cite{lin2025openvln,saxena2025uavvln,zhang2025logisticsvln,wu2025aeroduo,zhao2025aerialagent,yao2025aeroversereview}. Related aerial efforts, including aerial vision-dialog navigation and flying-on-a-word control, broaden the modality and deployment space under task assumptions that differ from sequential long-horizon instruction following~\cite{fan2023aerial,wang2025uavflow}. The common lesson is that methods inherited from ground VLN must be redesigned for aerial ambiguity, longer memory horizons, and more fragile sequential decisions.

\begin{figure*}[!t]
    \centering
    \includegraphics[width=0.86\textwidth]{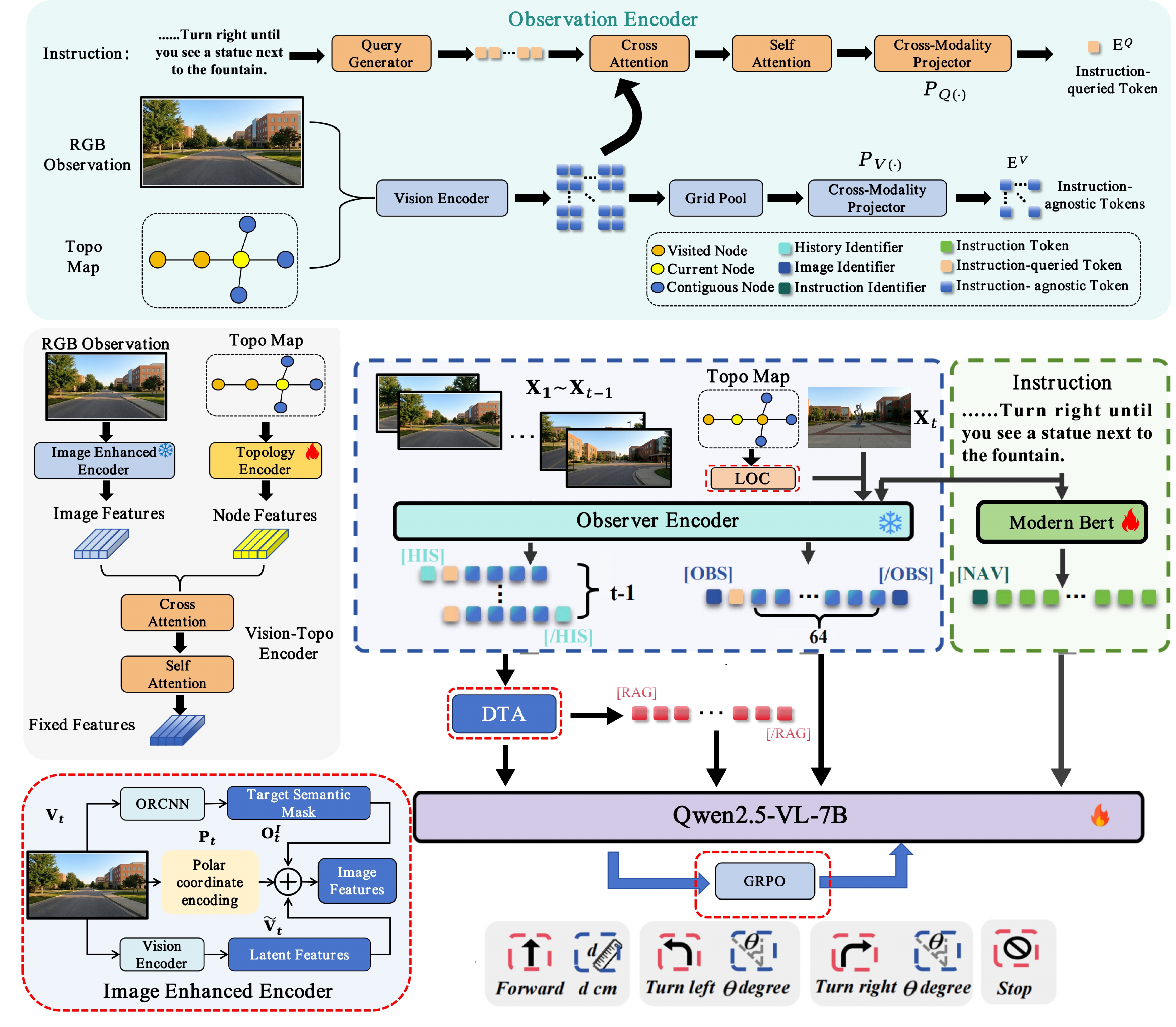}
    \caption{Overview of the proposed semantic-to-decision pipeline. The model encodes observation, instruction, and topology context; performs semantic-spatial enhancement and dynamic temporal aggregation (DTA) with a sparse grounding prompt branch, and conducts local optimum recovery (LOC) and GRPO-based decision optimization prior to next-action prediction.}
    \Description{A method overview for UAV vision-language navigation. Inputs on the left include the current RGB view, instruction tokens, history frames, and the local topology graph. The first stage builds an instruction-grounded current observation through a hybrid visual-topology encoder, fine-grained landmark semantics, and relative spatial encoding. The second stage performs weighted aggregation over all history frames, then applies sparse grounding to high-weight key frames to form a structured history prompt, yielding a filtered temporal state plus auxiliary prompt context. The third stage detects local stagnation on the evolving graph, re-ranks reachable frontier nodes, applies composite rewards, and refines the decoder policy with GRPO to produce the final action distribution.}
    \label{fig:framework}
\end{figure*}

\subsection{Semantic Grounding and History Modeling in Aerial Navigation}
The first two bottlenecks are fine-grained semantic grounding and selective history modeling. Small or similar entities, domain shifts, and limited labels challenge UAV detection~\cite{huang2022ufpmpdet,du2023ceasc,wu2024uavadapt,wu2026dronefine,li2026uavgen}, so coarse alignment can miss the landmark phrase that determines the next action. Ground VLN methods such as RCM, entity-graph reasoning, and HAMT improved token-level alignment and history use across VLN settings~\cite{wang2019reinforced,hong2020language,hao2020hamt}. More recent general-domain VLN work such as FlexVLN, SmartWay, NavMorph, and NavForesee also reinforces the need for cross-task adaptation, backtracking-aware reasoning, and more structured predictive use of navigation history~\cite{zhang2025flexvln,shi2025smartway,yao2025navmorph,liu2025navforesee}. Aerial studies expose the same failure mode from different angles: AerialVLN introduced recurrent aerial policies with look-ahead guidance, FELA strengthens fine-grained alignment in aerial vision-dialog navigation, STMR and CityNavAgent rely on stronger semantic reasoning, FlightGPT emphasizes interpretable VLM-based planning, and NaVid shows that video-based state construction can improve long-horizon generalization~\cite{liu2023aerialvln,su2025fela,gao2024stmr,zhang2025citynavagent,cai2025flightgpt,zhou2024navid}. A second challenge is that not all historical observations remain useful after a sub-goal has been completed. History-aware VLN models such as DUET show that selective memory matters in continuous navigation, and aerial systems such as OpenFly-Agent, SkyVLN, and FlySearch further support keyframe- or exploration-aware reasoning~\cite{chen2022think,gao2025openfly,li2025skyvln,pardyl2025flysearch}. Our design separates these roles explicitly: it first constructs instruction-grounded current-view tokens and only then applies relevance-aware temporal aggregation with a sparse grounding side branch.

\subsection{Topology-Aware Planning and Policy Optimization}
Even with stronger perception and memory, long-horizon UAV-VLN still fails when the decision state is not stable enough to avoid loops, false stopping, or semantically plausible detours. Topological reasoning is useful in VLN because it stabilizes long-range exploration, exposes frontier structure, and reduces local ambiguity~\cite{chen2021topological,qi2021orist}. In aerial navigation, STMR and CityNavAgent use structured spatial priors, GeoNav introduces dual-scale geospatial reasoning for coarse-to-fine aerial search, and Fly0 decouples semantic grounding from geometric planning~\cite{gao2024stmr,zhang2025citynavagent,xu2026geonav,xu2026fly0}. Adjacent map-based or control-oriented systems such as SkyVLN and UAV-Flow improve global consistency or low-level execution, but they reformulate the problem as location prediction, map search, or control rather than benchmark-style sequential policy learning~\cite{li2025skyvln,wang2025uavflow}. FlightGPT further explores GRPO-style post-training for aerial VLM agents, but its emphasis remains reasoning-process supervision rather than topology-conditioned state refinement~\cite{cai2025flightgpt}. Our aim is therefore narrower than generic aerial reasoning: we connect semantic grounding, relevance-aware temporal aggregation, and topology-aware decision optimization into one policy-state pipeline for sequential long-horizon UAV-VLN.

\section{The Proposed Approach}
\subsection{Problem Formulation}
We formulate UAV-VLN as an instruction-conditioned partially observable Markov decision process (POMDP) $\mathcal{M}=(\mathcal{S},\mathcal{A},P,\mathcal{O},\Omega,r,\gamma)$, where $P$ and $\Omega$ are the transition and observation models. An episode contains instruction $\mathcal{L}=\{w_1,\ldots,w_N\}$, initial latent state $s_1\in\mathcal S$, and hidden goal region $g$. At step $t$, the policy receives an egocentric RGB-D observation $o_t=(I_t,D_t)$ and an onboard pose estimation $\hat{\mathbf q}_t=(\hat x_t,\hat y_t,\hat z_t,\hat\psi_t)$ in an episode-local frame. It retains $\mathcal{H}_t=\{(o_\tau,\hat{\mathbf q}_\tau)\}_{\tau\in\mathcal B_t}$, where $\mathcal B_t=\{\tau\in\mathbb N\mid \max(1,t-H)\leq\tau<t\}$ with $\mathcal B_1=\varnothing$, and updates topology graph $\mathcal{G}_t=(\mathcal{V}_t,\mathcal{E}_t)$ from these estimated poses. Its policy information state is
\begin{equation}
x_t=(o_t,\hat{\mathbf q}_t,\mathcal{H}_t,\mathcal{G}_t,\mathcal{L}),
\end{equation}
where a rollout is $\xi=(x_1,a_1,\ldots,x_T,a_T,x_{T+1})$. The pose estimation $\hat{\mathbf q}_t$ is maintained from executed egomotion and the simulator's odometry-equivalent onboard stream (\emph{e.g.}, depth plus IMU/GPS/VIO) without goal coordinate. This sensor-side separation is consistent with RGB-D aerial benchmarks and realistic UAV sensor platforms~\cite{liu2023aerialvln,wang2025openuav}. In contrast, $s_t$ is the privileged simulator state used only by the transition, reward, and evaluation interfaces. The decoder context $c_t^{\mathrm{dec}}$ is constructed from the current observation encoder output, the DTA-filtered history representation, the structured prompt extracted from key historical frames, and the topology-aware target produced by LOC. The action policy is therefore written as $\pi_\theta(a_t\mid c_t^{\mathrm{dec}})$ over the discrete action space $\mathcal{A}$, which contains reachable motion choices and the ``\textsc{Stop}'' action. As the reward depends on post-action progress and stopping behavior, the instantiated step reward is $r_t=r(s_t,a_t,s_{t+1};g)$. With discounted return $R(\xi)=\sum_{t=1}^{T}\gamma^{t-1}r_t$, the training objective is
\begin{equation}
\max_{\pi_\theta}\; \mathbb{E}_{\xi\sim\pi_\theta}\!\left[R(\xi)\right],
\quad
a_t\sim\pi_\theta(\cdot\mid c_t^{\mathrm{dec}}),
\end{equation}
where the composite form of $r_t$ is defined in Sec.~\ref{sec:decision_opt}. Throughout the method, $s_t$, $\hat{\mathbf q}_t$, $x_t$, $\hat{f}_t$, $\tilde{h}_t$, $p_t^{\mathrm{rag}}$, and $v_t^\star$ denotes the latent simulator state, the policy-accessible local pose estimate, the policy information state, the instruction-enhanced current observation feature, the DTA-filtered history representation, the structured history prompt built from key historical frames, and the topology-aware frontier target selected by LOC, respectively. The resulting modeling problem is therefore how to transform $x_t$ into a decision context that remains semantically faithful to the instruction and topologically robust over long horizons.
We evaluate trajectory performance by three standard VLN metrics, including navigation error (NE), success rate (SR), and oracle success rate (OSR).

\subsection{Framework Overview}
Figure~\ref{fig:framework} summarizes the full pipeline: \emph{observation (vision + instruction) encoding} $\rightarrow$ \emph{fine-grained semantic and spatial enhancement} $\rightarrow$ \emph{dynamic temporal aggregation (DTA)} $\rightarrow$ \emph{local-optimum cognition (LOC)} $\rightarrow$ \emph{GRPO-based policy refinement}. For clarity, we group these operations into three macro stages. Stage~1 constructs an instruction-grounded current state from the RGB-D observation, the visual encoder, and the topology encoder. Stage~2 performs DTA over the retained history and augments the weighted main branch with a sparse grounding branch that converts up to two high-relevance frames into structured prompt memory. Stage~3 monitors the evolving topology graph, detects stagnation, and conditionally supplies a semantically aligned frontier cue to the decoder. GRPO then refines the trainable policy under the composite reward during training and is removed during inference. This decomposition preserves the explicit causal roles: current-view grounding determines what the agent should attend to at the current step, DTA filters relevant historical evidence, and LOC defines when topology informs the next action.

\subsection{Instruction-Grounded Perception and Temporal Modeling}
\paragraph{Observation (vision + instruction) encoder.}
The current-view encoder takes the RGB image $I_t$ as input and yields instruction-agnostic visual tokens, while a text encoder generates instruction tokens and an instruction query. Rather than being incorporated as an additional color channel, the aligned depth map $D_t$ is used by the spatial-cue branch. The topology graph is encoded in parallel and fused with the grounded observation following semantic-spatial enhancement. History, current-observation, and instruction streams remain distinguishable through the [HIS], [OBS], and [NAV] identifiers before being fed into the decoder. Specific backbone choices are described in Sec.~\ref{sec:implementation}.

\paragraph{Fine-grained semantic and spatial enhancement.}
Let $f_t^{\mathrm{base}}\in\mathbb{R}^{d_h}$ denote the base current-view token generated from RGB image $I_t$. In parallel, an object detector extracts $K_t$ landmark proposals $\{(b_t^k,v_t^k)\}_{k=1}^{K_t}$ from $I_t$, where $b_t^k$ is a landmark region and $v_t^k\in\mathbb{R}^{d_o}$ is its semantic descriptor. Let $\tilde d_t^k$ be the median valid value of aligned depth map $D_t$ within region $b_t^k$. Given camera calibration $\mathcal C$, we back-project the proposal center at depth $\tilde d_t^k$ and transform the resulting 3D point $\mathbf u_t^k$ into the UAV body frame. Defining $\bar{\mathbf u}_t^k=\mathbf u_t^k/d_{\mathrm{sens}}$, we encode the relative spatial relation as
\begin{equation}
p_t^k=\phi(b_t^k,D_t,\mathcal C)=
\left[\sin\vartheta_t^k,\cos\vartheta_t^k,
\bar\rho_t^k,\bar\zeta_t^k,\bar\alpha_t^k\right],
\label{eq:spatial_encode}
\end{equation}
where $\vartheta_t^k$ is the bearing angle in the body frame, $\bar\rho_t^k$, $\bar\zeta_t^k$, and $\bar\alpha_t^k$ are the clipped normalized range, vertical offset, and box scale, respectively. The box scale is $\sqrt{|b_t^k|/(WH)}$, where $W\times H$ is the image size, and spatial coordinates are normalized by sensor range $d_{\mathrm{sens}}$. Invalid-depth proposals use a learned unknown-spatial embedding. The resulting cue therefore integrates depth-derived bearing and relative 3D position with image-space box scale. The text encoder outputs the instruction-token matrix $H_{\mathcal L}^{\mathrm{tok}}\in\mathbb{R}^{N\times d_h}$ and a pooled instruction query $\bar e_{\mathcal L}\in\mathbb R^{d_h}$. Landmark relevance is modeled as
\begin{equation}
\eta_t^k = \mathrm{softmax}_k\left((W_q^{\mathrm{obj}} \bar{e}_{\mathcal{L}})^\top (W_k^{\mathrm{obj}}[v_t^k\|p_t^k])\right),
\end{equation}
and the instruction-grounded current-view feature becomes
\begin{equation}
\hat{f}_t = \mathrm{LN}\left(
W_g f_t^{\mathrm{base}} + \sum_{k=1}^{K_t}\eta_t^k W_o [v_t^k\|p_t^k]
\right).
\end{equation}
When no proposals are retained ($K_t=0$), the object residual is set to zero, yielding $\hat f_t=\mathrm{LN}(W_gf_t^{\mathrm{base}})$. This module injects fine-grained object-level semantics and relative position cues into the current observation, preventing the state from being dominated by coarse global appearance.

\paragraph{Visual-topological fusion.}
Long-horizon aerial navigation also depends on structural reachability. We encode the evolving graph to obtain node embeddings $\{u_t^m\}_{m=1}^{|\mathcal{V}_t|}$, stack them into matrix $U_t$, and compute the fused current-state token via cross-attention:
\begin{equation}
h_t = \mathrm{LN}\!\left(\hat{f}_t +
\mathrm{MHA}(Q=\hat{f}_t,K=U_t,V=U_t)\right).
\end{equation}
Here, $\hat f_t$ carries fine-grained semantic evidence from the current view, while $U_t$ encodes local reachability and neighborhood structure that cannot be recovered from appearance alone. Cross-attention highlights graph nodes that are both spatially adjacent and semantically aligned with the grounded instruction. The token $h_t$ is both instruction-grounded and topology-aware, preserving a clear interface to the successive decision modules.

\paragraph{Relevance-aware dynamic temporal aggregation.}
DTA aggregates all frames indexed by $\mathcal B_t$, while grounding only high-weight key frames. Taking $\bar e_{\mathcal L}$ as query, it computes the relevance weight for $\tau\in\mathcal B_t$ as
\begin{equation}
\alpha_{t,\tau} =
\frac{\exp\left((W_q^{\mathrm{hist}}\bar{e}_{\mathcal{L}})^\top W_h h_\tau / \sqrt{d_h}\right)}
{\sum_{j\in\mathcal B_t}\exp\left((W_q^{\mathrm{hist}}\bar{e}_{\mathcal{L}})^\top W_h h_j / \sqrt{d_h}\right)}.
\end{equation}
The main DTA branch combines the weighted full history buffer with the current observation to obtain the filtered temporal state
\begin{equation}
\tilde{h}_t = \mathrm{LN}\!\left(h_t+
\sum_{\tau\in\mathcal B_t}\alpha_{t,\tau}W_vh_\tau\right).
\end{equation}
This primary branch is the recurrent state pathway used by the navigation policy. In addition, DTA includes a sparse key-frame grounding branch to capture delayed landmark cues. We first select the high-weight historical index set
\begin{equation}
\mathcal{I}_t=\operatorname{TopK}_{\tau\in\mathcal B_t}
\left(\alpha_{t,\tau},\min(K_r,|\mathcal B_t|)\right),
\end{equation}
where $K_r=2$ in all reported experiments. For an empty initial buffer, the history sum and $m_t$ are set to zero with  $\mathcal I_t=\varnothing$. When the buffer contains only one frame, that single frame is selected. Otherwise, a frozen grounding model is applied to the selected RGB images, and its category--region outputs are paired with aligned depth through $\phi$:
\begin{equation}
\begin{aligned}
\{(\kappa_{\tau,m},b_{\tau,m})\}_{m=1}^{M_\tau}
&=\Gamma(I_\tau,\mathcal L),\\
\mathcal Z_\tau
&=\{(\kappa_{\tau,m},b_{\tau,m},
\phi(b_{\tau,m},D_\tau,\mathcal C))\}_{m=1}^{M_\tau},
\end{aligned}
\end{equation}
for $\tau\in\mathcal I_t$. Each tuple thereby encodes the landmark category, image region, and depth-derived relative position. These grounded tuples are serialized into a structured historical prompt
\begin{equation}
p_t^{\mathrm{rag}}=\operatorname{Prompt}\!\left(\{\mathcal Z_\tau\}_{\tau\in\mathcal{I}_t}\right),
\end{equation}
and further encoded as an auxiliary decoder memory
\begin{equation}
m_t=\operatorname{LMEnc}(p_t^{\mathrm{rag}}).
\end{equation}
\textsc{Prompt} also inserts a short descriptor of each selected frame, ensuring the decoder receives both frame-level context and landmark-level structure. The weighted aggregation branch outputs $\tilde h_t$ as the main temporal state, while the sparse grounding branch outputs $p_t^{\mathrm{rag}}$ and its encoded memory $m_t$ as auxiliary prompt context. Accordingly, DTA is not merely a top-$K$ retrieval module, but a weighted aggregation module augmented with a key-frame grounding side branch.

\begin{figure}[t]
    \centering
    \includegraphics[width=0.96\linewidth]{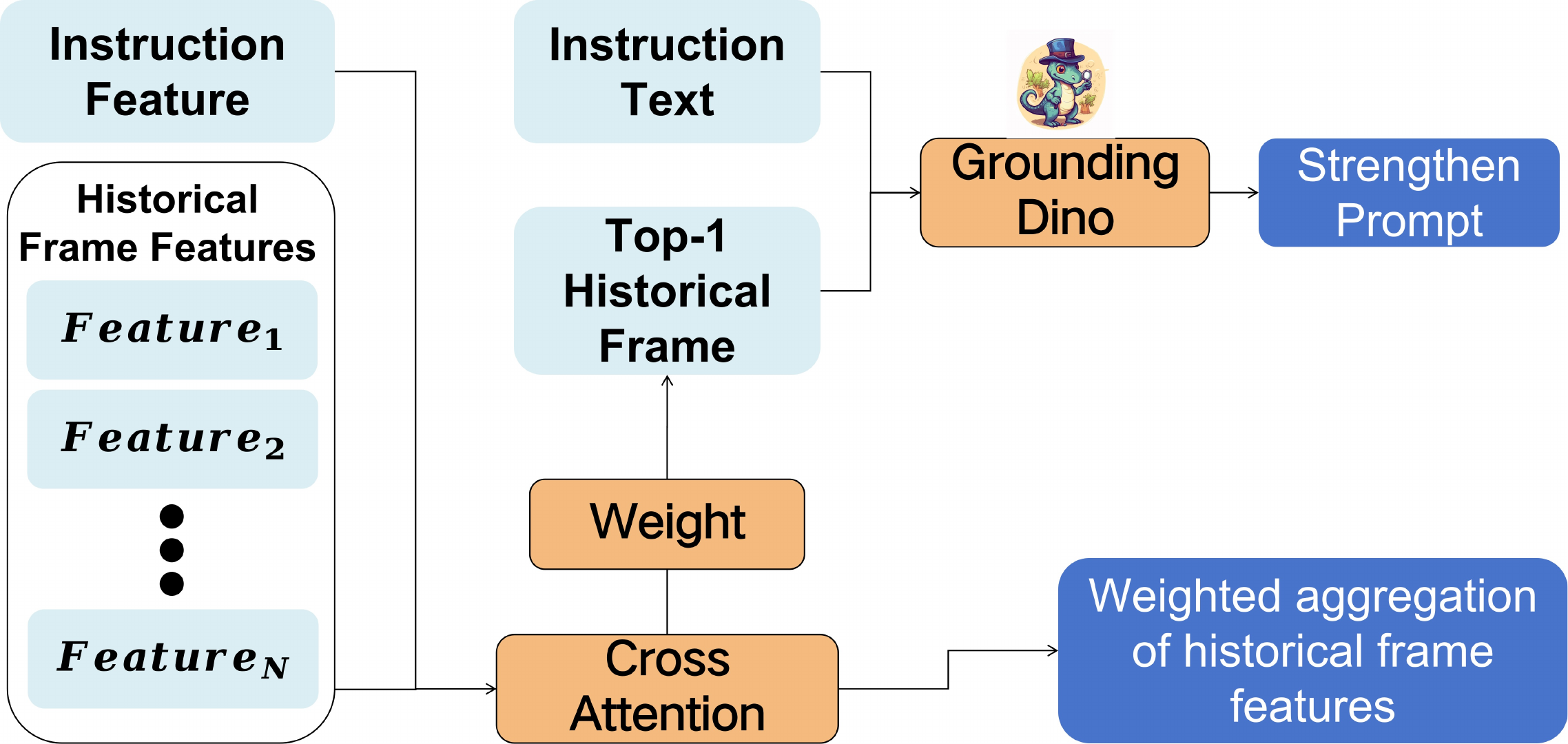}
    \caption{Overview of dynamic temporal aggregation with sparse grounding. The instruction query assigns relevance weights to historical observations to generate an aggregated history representation, while the top-$K$ frames are converted into structured landmark prompts with relative positional cues.}
    \Description{A method figure for dynamic temporal aggregation. The instruction feature assigns relevance weights over historical observations, forms an aggregated history representation, selects the top-$K$ most relevant frames for landmark-level structured prompting, and builds the final auxiliary prompt from frame descriptors, landmark JSON, and relative positional cues.}
    \label{fig:method_support}
\end{figure}

\subsection{Topology-Aware Decision Optimization}\label{sec:decision_opt}
Following current-view grounding and temporal filtering, the remaining challenge lies in decision robustness: even a well-refined semantic state can still induce local loops or semantically plausible yet ineffective detours. To mitigate this issue, we propose topology-aware local-optimum cognition (LOC) and GRPO-based policy optimization.

\paragraph{Local-optimum cognition.}
Let $\mathcal{V}_t^{\mathrm{exp}}\subseteq \mathcal{V}_t$ denote the explored nodes in the current dynamic topology graph, and let $\mathcal{F}_t\subset\mathcal{V}_t$ denote the reachable frontier nodes that remain unexplored. LOC monitors whether the exploration graph has expanded within the most recent $\Delta$ steps. For $t\ge\Delta$, stagnation is detected when no newly explored node appears during that window:
\begin{equation}
\mathbb{I}_{\mathrm{loc}}(t)=\mathbb{I}(t\ge\Delta)\,
\mathbb{I}\!\left(\mathcal{V}_t^{\mathrm{exp}}\setminus
\mathcal{V}_{t-\Delta}^{\mathrm{exp}}=\varnothing\right).
\end{equation}
For nonempty $\mathcal F_t$, let $e_v^{\mathrm{node}}\in\mathbb R^{d_h}$ and $v_t$ denote the semantic embedding of frontier node $v$, and the current graph node, respectively. LOC normalize semantic agreement and graph distance over $\mathcal F_t$. With distance extrema $d_t^{\min}$ and $d_t^{\max}$ and a constant $\epsilon_{\mathrm{loc}}>0$, it computes
\begin{equation}
\begin{aligned}
\bar s_t(v)
&=\frac{1+\cos(e_v^{\mathrm{node}},\bar e_{\mathcal L})}{2},\\
\bar d_{\mathcal G}(v_t,v)
&=\frac{d_{\mathcal G}(v_t,v)-d_t^{\min}}
{d_t^{\max}-d_t^{\min}+\epsilon_{\mathrm{loc}}},\\
v_t^\star
&=\arg\max_{v\in\mathcal F_t}
\left[\bar s_t(v)-\bar d_{\mathcal G}(v_t,v)\right].
\end{aligned}
\label{eq:loc_score}
\end{equation}
The two normalized terms represent instruction compatibility and relative traversal cost and are equally weighted. When LOC is active with a nonempty frontier set, the decoder cue encodes the selected node as $e_t^{\mathrm{loc}}=\mathrm{Enc}_{\mathrm{loc}}(v_t^\star)$; otherwise it is set to a zero vector.

\paragraph{Composite reward design.}
The transition reward balances four terms: progress reward, goal-completion reward, semantic matching reward, and path compliance penalty, which is defined as
\begin{equation}
r_t = r_t^{\mathrm{prog}} + r_t^{\mathrm{goal}} + r_t^{\mathrm{sem}} + r_t^{\mathrm{path}}.
\end{equation}
The goal region $g$ and benchmark-consistent distance $d(s,g)$ are available only to the simulator-side reward and evaluation modules, and are not included in $c_t^{\mathrm{dec}}$. To prevent distance measured in meters from dominating bounded semantic and terminal terms, we define the episode-normalized distance as $\bar d_t=\operatorname{clip}(d(s_t,g)/D_\xi,0,1)$, where $D_\xi=\max\{d(s_1,g),\epsilon_n\}$. The progress term rewards the agent for moving closer to the goal:
\begin{equation}
r_t^{\mathrm{prog}} =
\lambda_p \big(\bar d_t-\bar d_{t+1}\big).
\end{equation}
The goal-completion term assigns a terminal bonus when the agent reaches the goal region and executes the ``\textsc{Stop}'' action:
\begin{equation}
r_t^{\mathrm{goal}}=
\begin{cases}
\lambda_g, & a_t=``\textsc{Stop}"~\text{and}\ d(s_{t+1},g)\le \epsilon_g;\\
0, & \text{otherwise.}
\end{cases}
\end{equation}
The semantic matching term scores image-text consistency of the reached observation:
\begin{equation}
r_t^{\mathrm{sem}} =
\lambda_s \cos\!\big(\psi_I(I_{t+1}),\psi_T(\mathcal{L})\big).
\end{equation}
$\psi_I$ and $\psi_T$ are frozen CLIP image and text encoders used only for reward computation. The same frozen image encoder is reused for comparing observations in the descriptor space.
The path compliance term penalizes backtracking and persistent deviation from the goal. Let $t_w=\max\{1,t-w\}$ denote the first index in the revisit window:
\begin{equation}
\begin{aligned}
r_t^{\mathrm{path}}={}&-\lambda_r \sum_{j=t_w}^{t-1}
\exp\!\big(-\mu(t-j)\big)\,
\mathbb{I}_{\mathrm{rev}}(t+1,j)\\
&-\lambda_c \, \bar n_{t+1}^{\mathrm{dev}} \, \bar d_{t+1},
\end{aligned}
\end{equation}
\noindent where the deviation counter is initialized as $n_1^{\mathrm{dev}}=0$ and evolves as
\begin{equation}
\begin{aligned}
n_{t+1}^{\mathrm{dev}}&=
\begin{cases}
n_t^{\mathrm{dev}}+1, & d(s_{t+1},g)>d(s_t,g)+\epsilon_d,\\
0, & \text{otherwise,}
\end{cases}\\
\bar n_{t+1}^{\mathrm{dev}}&=\min\{n_{t+1}^{\mathrm{dev}},w\}/w.
\end{aligned}
\end{equation}
and the revisit indicator is defined as
\begin{equation}
\begin{aligned}
s_{\mathrm{vis}}(t,j)
&=
\frac{1+\cos\!\big(\psi_I(I_t),\psi_I(I_j)\big)}{2},\\
\mathbb{I}_{\mathrm{rev}}(t,j)
&=
\mathbb{I}\!\left(d_{3\mathrm{D}}(\hat{\mathbf q}_t,\hat{\mathbf q}_j)\le \delta_r\right)\\
&\quad\cdot
\mathbb{I}\!\left(|\hat z_t-\hat z_j|\le \delta_h\right)\\
&\quad\cdot
\mathbb{I}\!\left(s_{\mathrm{vis}}(t,j)\ge \rho_v\right).
\end{aligned}
\end{equation}
Here $d_{3\mathrm D}$ uses only the positional components of the estimated local poses, $\hat z_t$ is the altitude component, $\epsilon_g$ is the success radius, and $s_{\mathrm{vis}}(t,j)\in[0,1]$ is the normalized cosine similarity between frozen global image descriptors. This descriptor-based verification compares observations in a learned embedding space, eliminating the need for pixel-coordinate alignment of detection boxes across different camera views. The joint estimated-spatial, altitude, and visual tests suppress false revisit matches caused by nearby but visually distinct locations. Since $\bar d_t,\bar n_t^{\mathrm{dev}}\in[0,1]$, both progress and persistent-deviation terms are bounded independently of route length and metric scale.

\paragraph{GRPO-based policy refinement.}
The decoder context for action prediction is assembled as
\begin{equation}
c_t^{\mathrm{dec}}=\operatorname{Fuse}_{\mathrm{dec}}\!\left(
\tilde{h}_t,\,
m_t,\,
e_t^{\mathrm{loc}},\,
H_{\mathcal L}^{\mathrm{tok}}
\right),
\end{equation}
where $H_{\mathcal L}^{\mathrm{tok}}$ is the instruction-token matrix defined above, and the remaining inputs are the DTA state, structured-prompt memory, and conditional LOC cue. Instead of training a separate critic network, we adopt GRPO for policy refinement over this decoder context~\cite{schulman2017ppo,shao2024deepseekmath,cai2025flightgpt}. For each instruction and start state, we sample a group of $M$ trajectories $\{\xi_i\}_{i=1}^{M}$ from $\pi_{\theta_{\mathrm{old}}}$, and compute group-normalized advantages as
\begin{equation}
\begin{aligned}
A_i
&=
\frac{R(\xi_i)-\mu_R}{\sigma_R+\epsilon_A},\\
\mu_R
&=
\frac{1}{M}\sum_{i=1}^{M}R(\xi_i),\\
\sigma_R^2
&=
\frac{1}{M}\sum_{i=1}^{M}\left(R(\xi_i)-\mu_R\right)^2,
\end{aligned}
\end{equation}
where $\epsilon_A>0$ handles identical group returns. The clipped GRPO surrogate is defined as
\begin{equation}
\begin{aligned}
\mathcal{J}_{\mathrm{GRPO}}(\theta)
&=
\mathbb{E}\Bigg[
\frac{1}{M}\sum_{i=1}^{M}\frac{1}{T_i}
\sum_{t=1}^{T_i}
\left(
\ell_{i,t}(\theta)
-\beta D_{i,t}^{\mathrm{KL}}(\theta)
\right)
\Bigg],\\
\ell_{i,t}(\theta)
&=
\min\Bigg(
\rho_{i,t}(\theta)A_i,\,
\mathrm{clip}\big(\rho_{i,t}(\theta),1-\epsilon_c,1+\epsilon_c\big)A_i
\Bigg),
\end{aligned}
\end{equation}
where
\begin{equation}
\begin{aligned}
\rho_{i,t}(\theta)
&=
\frac{\pi_\theta(a_{i,t}\mid c_{i,t}^{\mathrm{dec}})}
{\pi_{\theta_{\mathrm{old}}}(a_{i,t}\mid c_{i,t}^{\mathrm{dec}})},\\
D_{i,t}^{\mathrm{KL}}(\theta)
&=
D_{\mathrm{KL}}\!\left(
\pi_\theta(\cdot\mid c_{i,t}^{\mathrm{dec}})
\;\middle\|\;
\pi_{\mathrm{ref}}(\cdot\mid c_{i,t}^{\mathrm{dec}})
\right).
\end{aligned}
\end{equation}
Here $T_i$ is the trajectory length. The old policy $\pi_{\theta_{\mathrm{old}}}$ is the rollout snapshot used only in the importance ratio and is refreshed between update rounds, whereas $\pi_{\mathrm{ref}}$ is a frozen copy of the behavior-cloned policy at the start of GRPO fine-tuning. The per-decision KL divergence term anchors the learned policy to the supervised initialization, while clipping costrains each update relative to the rollout policy. Maximizing $\mathcal J_{\mathrm{GRPO}}$ couples policy refinement to the DTA state, sparse prompt memory, and LOC cue.

\begin{table*}[!t]
\caption{Comparison of performance on the validation seen and unseen splits of AerialVLN-S. The best results are highlighted in bold and the second-best ones are underlined.}
\label{tab:aerial_main}
\centering
\small
\setlength{\tabcolsep}{4.0pt}
\renewcommand{\arraystretch}{0.88}
\begin{tabular}{lcccccc}
\toprule
\multirow{2}{*}{Method} & \multicolumn{3}{c}{Validation Seen} & \multicolumn{3}{c}{Validation Unseen} \\
\cmidrule(lr){2-4}\cmidrule(lr){5-7}
& NE$\downarrow$ & SR$\uparrow$ & OSR$\uparrow$ & NE$\downarrow$ & SR$\uparrow$ & OSR$\uparrow$ \\
\midrule
Random & 109.60 & 0.00 & 0.00 & 149.70 & 0.00 & 0.00 \\
Action Sampling & 213.80 & 0.90 & 5.70 & 237.60 & 0.20 & 1.10 \\
Seq2Seq~\cite{anderson2018vision} & 146.00 & 4.80 & 19.80 & 218.90 & 2.30 & 11.70 \\
CMA~\cite{liu2023aerialvln} & 121.00 & 3.00 & 23.20 & 172.10 & 3.20 & 16.00 \\
LAG~\cite{liu2023aerialvln} & 90.20 & 7.20 & 15.70 & 127.90 & 5.10 & 10.50 \\
STMR~\cite{gao2024stmr} & 63.82 & 30.15 & 56.30 & 97.86 & 18.35 & 33.40 \\
CityNavAgent~\cite{zhang2025citynavagent} & 66.34 & 28.97 & 58.10 & 100.31 & 16.70 & 31.90 \\
NavGPT~\cite{zhou2024navgpt} & 59.41 & 34.04 & 61.20 & 91.60 & 26.50 & 46.20 \\
SPF~\cite{hu2025spf} & 40.12 & 45.89 & 72.38 & 71.34 & 35.12 & 58.30 \\
OpenFly-Agent~\cite{gao2025openfly} & 66.93 & 36.98 & 61.80 & 90.67 & 29.31 & 47.80 \\
UAV-Flow~\cite{wang2025uavflow} & 58.16 & 35.81 & 58.34 & 82.38 & 28.85 & 46.10 \\
Fly0~\cite{xu2026fly0} & \textbf{27.19} & \underline{70.43} & \underline{81.20} & \underline{51.23} & \underline{60.07} & \underline{69.22} \\
FlightGPT~\cite{cai2025flightgpt} & 38.17 & 44.62 & 72.98 & 62.88 & 31.75 & 60.09 \\
\midrule
\textbf{Ours} & \underline{28.34} & \textbf{71.12} & \textbf{82.60} & \textbf{50.69} & \textbf{61.38} & \textbf{71.01} \\
\bottomrule
\end{tabular}
\end{table*}

\begin{table}[!t]
    \caption{Comparison of performance on the test set of OpenFly. The best results are highlighted in bold and the second-best ones are underlined.}
\label{tab:openfly_main}
\centering
    \small
    \setlength{\tabcolsep}{4.0pt}
    \renewcommand{\arraystretch}{0.88}
\begin{tabular}{lccc}
\toprule
Method & NE$\downarrow$ & SR$\uparrow$ & OSR$\uparrow$  \\
\midrule
Random & 165.38 & 0.00 & 0.00 \\
OpenUAV~\cite{wang2025openuav} & 74.63 & 17.81 & 31.16 \\
UAV-Flow~\cite{wang2025uavflow} & 69.51 & 32.14 & 44.01 \\
AerialVLN~\cite{liu2023aerialvln} & 94.63 & 6.71 & 19.91 \\
OpenFly~\cite{gao2025openfly} & 64.08 & 33.61 & 49.72 \\
NavGPT~\cite{zhou2024navgpt} & 61.57 & 29.28 & 43.48 \\
STMR~\cite{gao2024stmr} & 66.54 & 27.43 & 42.64 \\
CityNavAgent~\cite{zhang2025citynavagent} & 70.24 & 25.33 & 39.75 \\
SPF~\cite{hu2025spf} & 46.57 & 42.94 & 55.79 \\
Fly0~\cite{xu2026fly0} & \underline{29.47} & \underline{64.67} & \underline{72.81} \\
FlightGPT~\cite{cai2025flightgpt} & 41.29 & 56.23 & 69.03 \\
\midrule
\textbf{Ours} & \textbf{26.14} & \textbf{67.31} & \textbf{77.32} \\
\bottomrule
\end{tabular}
    \renewcommand{\arraystretch}{1.0}
\end{table}

\subsection{Training and Inference}
For training, we first initialize the decoder with behavior cloning on expert trajectories, ensuring the policy starts GRPO fine-tuning from a stable instruction-following regime:
\begin{equation}
\mathcal{L}_{\mathrm{BC}} = - \sum_t \log \pi_\theta(a_t^\star \mid c_t^{\mathrm{dec}}),
\end{equation}
where $a_t^\star$ is the expert action and $c_t^{\mathrm{dec}}$ follows the decoder interface above. We then fine-tune the trainable policy modules with GRPO, yielding
\begin{equation}
\mathcal{L}_{\mathrm{train}}=
\begin{cases}
\mathcal{L}_{\mathrm{BC}}, & \text{stage 1},\\
-\mathcal{J}_{\mathrm{GRPO}}, & \text{stage 2}.
\end{cases}
\end{equation}
During inference, the system first encodes the current RGB image and instruction, derives depth-based semantic-spatial cues, and fuses the grounded observation with the local topology to obtain $h_t$. DTA then weights the retained history to generate the main temporal state $\tilde h_t$, while its sparse branch converts up to two high-relevance historical RGB-D frames into the structured prompt $p_t^{\mathrm{rag}}$ and auxiliary memory $m_t$. When LOC detects stagnation and a reachable frontier exists, it conditionally appends the selected frontier cue $e_t^{\mathrm{loc}}$ to the decoder context. The decoder predicts the next action from $c_t^{\mathrm{dec}}$. GRPO group sampling and policy updates are used only during training and are not part of the inference path.

\section{Experimental Results and Analysis}
\subsection{Datasets and Evaluation Metrics}
\textbf{AerialVLN.} AerialVLN~\cite{liu2023aerialvln} contains city-scale continuous UAV trajectories with long natural-language instructions and is one of the pioneering  UAV-VLN benchmarks. We report AerialVLN-S validation performance on the public seen and unseen splits with NE, SR, and OSR.

\textbf{OpenFly.} OpenFly~\cite{gao2025openfly} is a large-scale aerial VLN benchmark comprising 100K trajectories across 18 scenes rendered from multiple engines and representations. Our model is trained on the OpenFly training set and  evaluated on the test set under the benchmark's standard evaluation protocol.

We report three standard metrics: navigation error (NE) measuring the terminal straight-line distance to the target; success rate (SR) denoting the fraction of episodes stopping within 20 m of the target, and oracle success rate (OSR) denoting the fraction entering that radius at any trajectory step.

\subsection{Implementation Details}\label{sec:implementation}
We utilize Oriented R-CNN~\cite{xie2021orientedrcnn} through MMRotate~\cite{zhou2022mmrotate} for current-view enhancement, Grounding DINO~\cite{liu2024groundingdino} for sparse history grounding, Qwen2.5-VL-7B~\cite{bai2025qwen25vl} with LoRA~\cite{hu2022lora} as the action decoder, frozen CLIP~\cite{radford2021clip} for reward and revisit descriptors, and a lightweight GCN~\cite{kipf2017gcn} for topology modeling. All the main results are averaged over three random seeds. The complete geometry, history, training, reward, revisit, parameter-count, and efficiency configurations are provided in the supplementary material.

\subsection{Comparison with Strong Baselines}
\textbf{On AerialVLN-S validation.} The comparison results in Table~\ref{tab:aerial_main} show a clear progression from early recurrent baselines (\emph{e.g.},  Seq2Seq, CMA) to SOTA aerial reasoning systems, indicating that long-horizon UAV-VLN depends on accurate landmark grounding, effective history modeling, and stable sequential decisions. Our proposed framework achieves the highest SR on both validation splits and the highest OSR on the seen split, while remaining competitive in terms of NE.

\begin{figure}[!t]
    \centering
    \includegraphics[width=0.95\linewidth]{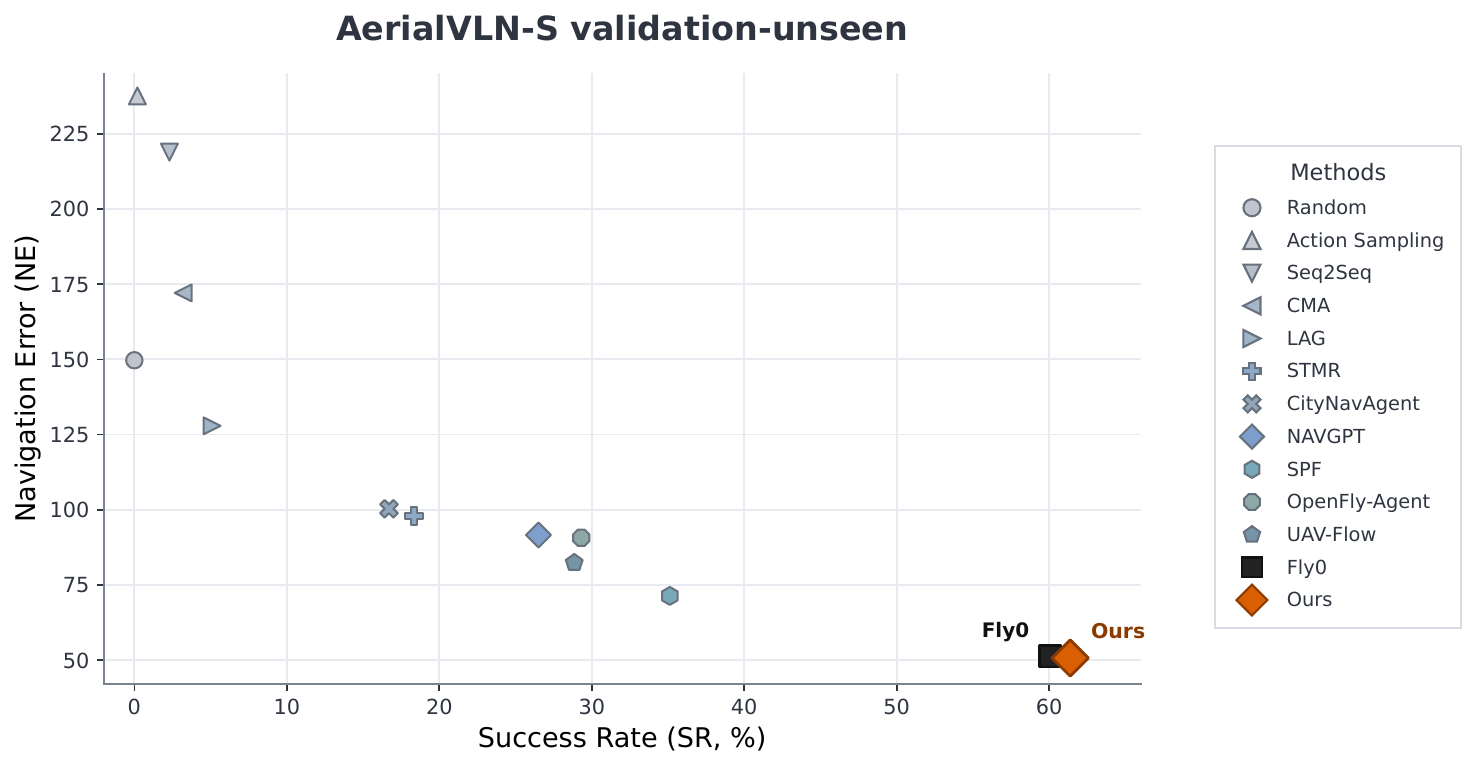}
    \caption{SR--NE comparison on the AerialVLN-S validation-unseen split. Each marker denotes one method; higher SR and lower NE indicate better performance. Our full model is highlighted for clarity.}
    \Description{A single-column scatter plot for AerialVLN-S validation-unseen with success rate on the horizontal axis and navigation error on the vertical axis, where lower error appears higher because the y-axis is inverted. Each method appears as one marker with a legend placed outside the plot on the right. Fly0 is highlighted with a dark square, and the final model Ours is highlighted with the largest orange diamond slightly above and to the right of Fly0.}
    \label{fig:trend}
\end{figure}

The unseen split serves as a rigorous stress test, as distribution shift amplifies landmark ambiguity and long-horizon error accumulation. Compared to Fly0~\cite{xu2026fly0}, our method improves unseen SR from 60.07 to 61.38 and reduces NE from 51.23 to 50.69. The three-seed results reliably quantify  run-to-run variance for our method, but they do not constitute a paired significance test against baseline values obtained from separate reports. The SR–NE trade-off on the unseen split is visualized in Figure~\ref{fig:trend}.

\textbf{On OpenFly.}
Our model is trained on the official training split on OpenFly and evaluated on the test set under the standard protocol. As summarized in Table~\ref{tab:openfly_main}, our method achieves the lowest NE and the highest SR and OSR among all reported methods.

Compared with Fly0~\cite{xu2026fly0}, our method reduces NE from 29.47 to 26.14, and improves SR from 64.67 to 67.31, and raises OSR from 72.81 to 77.32. The more pronounced performance gain may be attributed to  its fine-grained landmark and route cues.

Additional three-seed average results on the AerialVLN-S test split (53.5 NE, 59.0 SR, and 69.0 OSR) and the OpenFly validation split (25.0 NE, 68.5 SR, and 78.5 OSR) are detailed in the supplementary material.

\subsection{Ablation Study}
Table~\ref{tab:ablation} reports fixed-order cumulative additions on the AerialVLN-S validation-unseen split. The consistent monotonic improvements across NE, SR, and OSR gains clearly validate the effectiveness of our coupled pipeline design.

\begin{table}[t]
    \caption{Cumulative ablation on the AerialVLN-S validation-unseen split. Each row adds one component in pipeline order to the same base model. The last row corresponds to our full model and matches the validation-unseen result reported in Table~\ref{tab:aerial_main}.}
\label{tab:ablation}
\centering
\small
\renewcommand{\arraystretch}{0.88}
\begin{tabular}{lccc}
\toprule
Variant & NE$\downarrow$ & SR$\uparrow$ & OSR$\uparrow$ \\
\midrule
Baseline & 72.10 & 35.85 & 59.92 \\
+ Semantic enhancement & 67.84 & 40.96 & 62.88 \\
+ DTA & 62.11 & 46.74 & 66.41 \\
+ Sparse grounding & 58.27 & 51.32 & 68.45 \\
+ LOC & 54.06 & 56.47 & 69.94 \\
+ GRPO (Full) & \textbf{50.69} & \textbf{61.38} & \textbf{71.01} \\
\bottomrule
\end{tabular}
\renewcommand{\arraystretch}{1.0}
\end{table}

Additional leave-one-out and alternative history results achieve SR of 46.74 without DTA, 50.90 without LOC, and 55.43 with simple history aggregation, compared to 61.38 for the full model. Reward ablations demonstrate consistent gains over Basic GRPO (57.42 SR): semantic-only, progress-only, goal/stop-only, and path-compliance-only obtain 58.52, 59.52, 58.24, and 58.91 SR, respectively, while the full composite reaches 61.38. Under Gaussian image noise with $\sigma=0.10$, our model yields 58.40 NE, 55.10 SR, and 64.85 OSR. Under moderate accumulated localization drift, our method reaches  56.71 NE, 56.98 SR, and 66.23 OSR. Full protocols, metrics, and evaluation boundaries are summarized in the supplementary material.

\subsection{Qualitative Analysis}
\begin{figure}[t]
    \centering
    \includegraphics[width=0.94\linewidth]{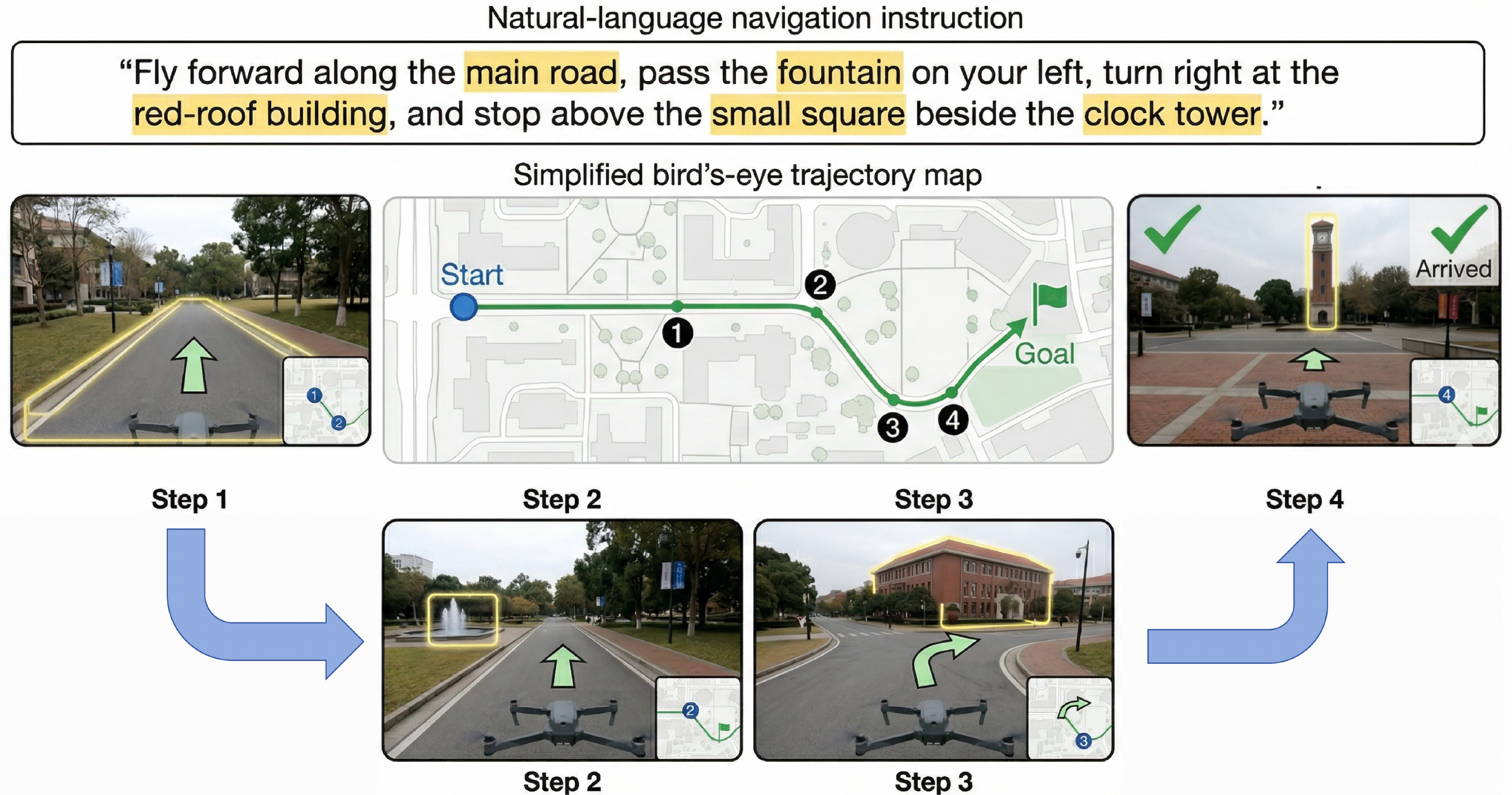}
    \caption{Qualitative example of a successful navigation episode  by the proposed method. It displays the instruction, four egocentric UAV observations, and the corresponding bird’s-eye trajectory. The agent follows the specified landmark order and terminates near the target region.}
    \Description{A qualitative UAV navigation figure showing one successful route. The instruction asks the agent to follow the main road, pass a fountain on the left, turn right at a red-roof building, and stop above a small square beside a clock tower. Four first-person UAV views and a bird's-eye map show the trajectory from Start to Goal through these landmarks.}
    \label{fig:qualitative}
\end{figure}

Figure~\ref{fig:qualitative} shows a representative successful case. The agent follows the main road, passes the fountain on its left, turns right at the red-roof building, and stops above the small square beside the clock tower. The trajectory follows the landmark order described in the instruction, illustrating semantically coherent long-horizon execution.

\subsection{Discussion}
The experimental results support the perspective that treats long-horizon UAV-VLN as a policy-state construction problem, rather than as isolated perception, memory, and planning subproblems. Results across AerialVLN-S and OpenFly, together with the ablation studies, consistently support instruction-grounded observations, selective history, and topology cues as a coupled policy state. These results should be interpreted within the evaluation design: three-seed statistics quantify run-to-run stability, Table~\ref{tab:ablation} reports cumulative additions, and supplementary experiments further validate the effect of  DTA and LOC. Dependence on external perception, onboard pose, and lightweight topology motivates further deployment-oriented validation beyond benchmark navigation.

\section{Conclusion}
This work presents a unified semantic-to-decision framework that connects current-view grounding, relevance-aware temporal aggregation, and topology-aware decision refinement for long-horizon UAV-VLN. DTA preserves retained history and recovers delayed semantic cues through sparse landmark prompts. And  LOC conditionally supplies a frontier cue, when the topology indicates stagnation. Extensive experiments on AerialVLN-S and OpenFly, together with component, reward, simulated image-noise, and localization-drift analyses, validate that these modules function as a coupled state construction process, enabling more stable long-horizon navigation behavior.  For future work, we will evaluate our framework under more challenging real-world conditions, including adverse weather, dynamic obstacles, stronger and sensor-coupled perception/localization corruption, and real closed-loop flight while exploring richer semantic-geometric memory and broader cross-environment validation.

\begin{acks}
This work was supported in part by the Beijing Natural Science Foundation (Grants 4242044 and L259044), the Research Program of the State Key Laboratory of Virtual Reality Technology and Systems, and the Fundamental Research Funds for the Central Universities.
\end{acks}
\begingroup
\renewcommand{\href}[2]{\unskip}
\providecommand{\showURL}[1]{\unskip}
\renewcommand{\showURL}[1]{\unskip}
\bibliographystyle{ACM-Reference-Format}
\bibliography{sample-base}
\endgroup

\end{document}